\documentclass[10pt,twocolumn,letterpaper]{article}

\usepackage{cvpr}              
\usepackage{graphicx}
\usepackage{subcaption}
\usepackage{multirow}
\usepackage{soul}
\usepackage{amsthm}
\usepackage{marvosym}
\usepackage{listings}
\definecolor{parchment}{RGB}{255, 239, 213}
\definecolor{cvprblue}{rgb}{0.21,0.49,0.74}
\usepackage[pagebackref,breaklinks,colorlinks,allcolors=cvprblue]{hyperref}

\newcommand{\blankfootnote}[1]{%
  \let\temp\thefootnote
  \renewcommand{\thefootnote}{}
  \footnotetext{#1}
  \let\thefootnote\temp
}

\def\paperID{4774} 

\def\confName{CVPR}
\def\confYear{2025}

\title{MiLA: \underline{M}ulti-view \underline{I}ntensive-fidelity \underline{L}ong-term Video Generation World Model for \underline{A}utonomous Driving }

\author{
Haiguang Wang$^{1,2,*,\Diamond}$\hspace{3mm}
Daqi Liu$^{2,*}$\hspace{3mm}
Hongwei Xie$^{2,\dag}$\hspace{3mm}
Haisong Liu$^{1}$\hspace{3mm}
Enhui Ma$^{3}$\hspace{3mm} 
\\
Kaicheng Yu$^{3}$\hspace{3mm}
Limin Wang$^{1,}$\textsuperscript{\Letter} \hspace{3mm}
Bing Wang$^{2}$\hspace{3mm}
\\ 
${}^{1}$Nanjing University \hspace{6mm}
${}^{2}$Xiaomi EV \hspace{6mm}
${}^{3}$Westlake University
\\
\texttt{\{haiguangwang, liuhs\}@smail.nju.edu.cn}\texttt{,}\\
\texttt{\{liudaqikk, hongwei.xie.90, Kaicheng.yu.yt, blucewang6\}@gmail.com}\texttt{,}\\
\texttt{maenhui@westlake.edu.cn}\texttt{,} \texttt{lmwang@nju.edu.cn}
}

\begin{document}

\twocolumn[{
  \maketitle
  \vspace{-10mm}
  \begin{center}
    \includegraphics[width=\textwidth]{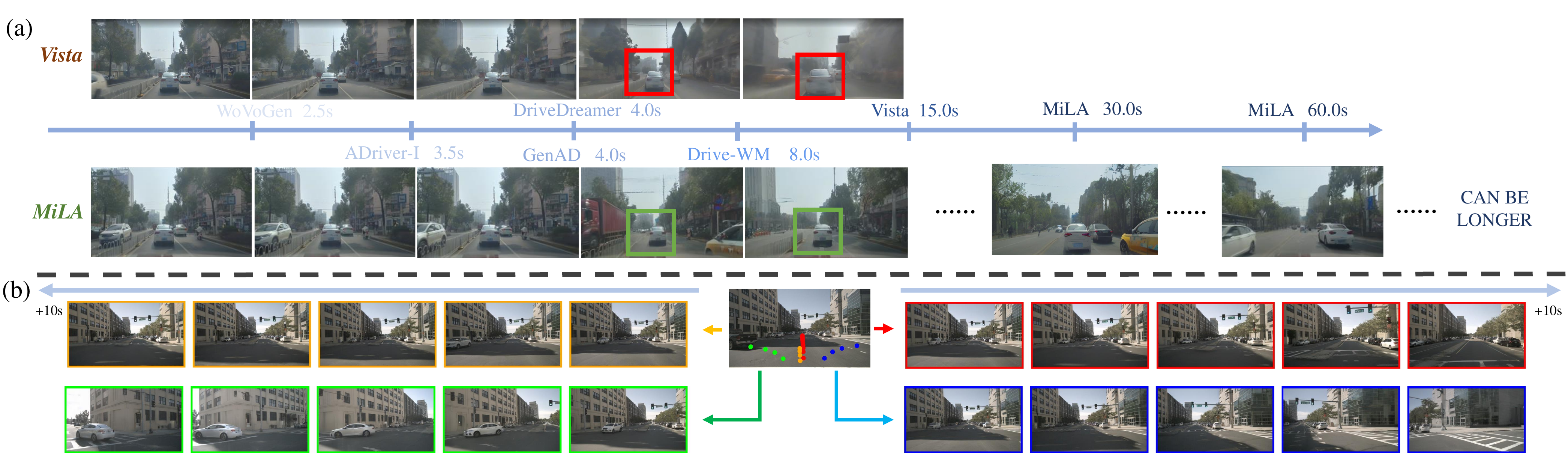}  
    \captionsetup{type=figure}  
    \vspace{-5mm}
    
    \caption{(a) Comparison between Vista and the proposed MiLA for long video generation. \textbf{\textit{Top: }}The generation frames of Vista. \textbf{\textit{Bottom: }}The generation frames of our MiLA. \textbf{\textit{Middle: }} A timeline in increasing order, where the value in second indicates the capability in generation length of other approaches, and the temporal time for generated frames. As seen in the red boxes, the performance of Vista significantly degrades after 8 seconds. In contrast, the definition of our MiLA remains largely unaffected, as highlighted in the green boxes. Furthermore, the 30-second frame predicted by MiLA maintains acceptable clarity throughout. (b) Visualization of videos generated by MiLA given different waypoints based on identical initial frames. }
    \label{fig:first_img}
  \end{center}
  \vspace{2mm}  
}]

\blankfootnote{
 $^{*}$ Equal contribution.
$^{\Diamond}$ Work done while an intern at Xiaomi EV.

\quad  $^{\dag}$ Project leader. \textsuperscript{\Letter} Corresponding author.

 
}

\begin{abstract}
In recent years, data-driven techniques have greatly advanced autonomous driving systems, but the need for rare and diverse training data remains a challenge, requiring significant investment in equipment and labor. World models, which predict and generate future environmental states, offer a promising solution by synthesizing annotated video data for training. However, existing methods struggle to generate long, consistent videos without accumulating errors, especially in dynamic scenes. To address this, we propose MiLA, a novel framework for generating high-fidelity, long-duration videos up to one minute. MiLA utilizes a Coarse-to-Re(fine) approach to both stabilize video generation and correct distortion of dynamic objects. Additionally, we introduce a Temporal Progressive Denoising Scheduler and Joint Denoising and Correcting Flow modules to improve the quality of generated videos. Extensive experiments on the nuScenes dataset show that MiLA achieves state-of-the-art performance in video generation quality. For more information, visit the project website: \url{https://github.com/xiaomi-mlab/mila.github.io}.
\end{abstract}   
\vspace{-12mm}

\section{Introduction}
\label{sec:intro}
\begin{figure*}[h!]
    \centering
    \includegraphics[width=1.0\linewidth]{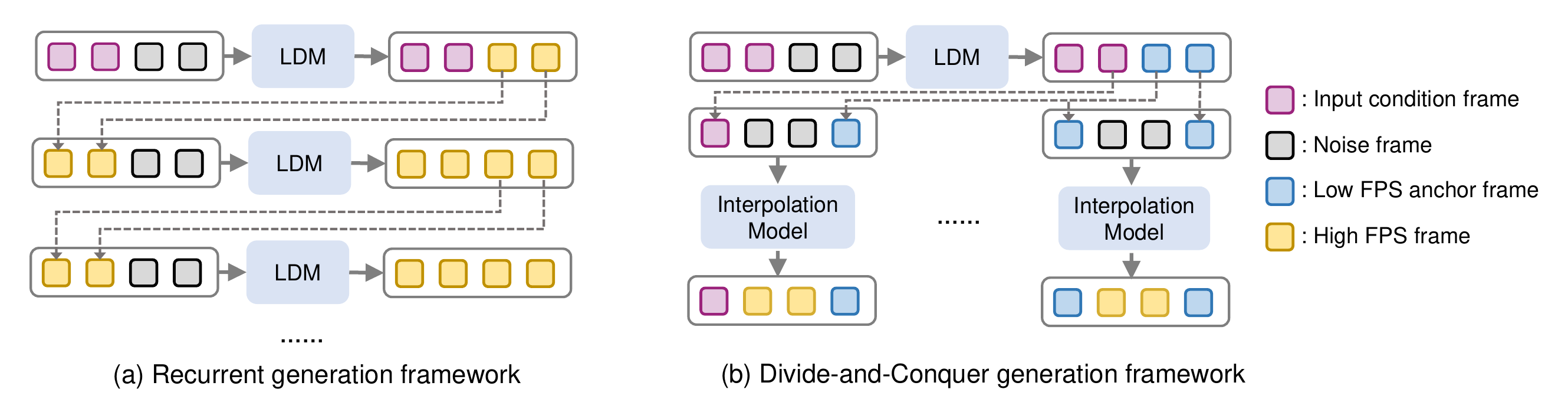}
    \caption{An illustration of the existing Recurrent and the existing Divide-and-Conquer generation frameworks. \textbf{\textit{Left:}} The generation of newly predicted frames always conditioned the previously generated frames recursively with an LDM~\cite{vista,Gaia-1}. \textbf{\textit{Right:}} Anchor frames are firstly drawn by an LDM, and frames between anchor frames are interpolated in parallel with an Interpolation Model~\cite{videoldm,nuwa}.}
    \label{fig:comparison_RVD}
    \vspace{-7mm}
\end{figure*}

Data-driven approaches have demonstrated remarkable success in full-autonomous driving systems~\cite{uniad,e2e_challenge,vad,genad}. However, building robust systems requires the collection of long-tail and rare scenario data, demanding substantial ongoing investments in equipment and human resources. In response, world model systems~\cite{vista,wovogen,Gaia-1,drivedreamer,drivedreamer2,occworld,vidar,delphi,drivewm} capable of understanding and predicting future environmental states have emerged as a promising solution. By processing historical observations and human instructions to generate scene representations, which can synthesize consistent sequences, effectively addressing out-of-distribution challenges~\cite{ad-ood-review,drive-anywhere}.

Recent years have witnessed significant progress in world model frameworks capable of generating high-quality videos with planning signals\cite{drivedreamer,drivedreamer2,drivewm, adriver}. However, two significant challenges persist in generating consistent long-duration data. The first challenge is the accumulated error problem. While generating all frames simultaneously would be ideal, it demands impractical computational resources and memory requirements. Consequently, most approaches adopt a batch-by-batch strategy for long video generation \cite{vista,wovogen,synthesizing_coherent,grimm-open-ended}, predicting new frames recursively based on previously generated ones. This approach, unfortunately, leads to error accumulation throughout the generation process, degrading the quality of both static backgrounds and dynamic objects in long-term videos. 


The second challenge lies in maintaining scene consistency over extended sequences. When conditioned only on the information with the initial timestamps, including conditional frames and scene descriptions, the control influence of the first frame gradually diminishes over subsequent frames, often resulting in scene degradation and loss of coherence in later portions of the sequence. While existing approaches have attempted to address this by incorporating extra control signals~\cite{delphi,tex4d,magicdrive3d,mygo}, such as HD-maps and 3D bounding boxes of the predicted frames. It remains largely unexplored to learn a representation without these signals while still maintaining spatial-temporal consistency.

To address these limitations, we present MiLA, a robust video generation framework built upon Latent Diffusion Models (LDM)~\cite{ldm,svd} capable of producing high-fidelity long-term videos. Precisely, our approach focuses on maintaining long-term consistency and stability while relying solely on simple scene descriptions or waypoints as control signals. Comparison with approaches that utilized the existing Recurrent framework and demonstration of waypoints controllability are shown in Fig.~\ref{fig:first_img}. We explore a divide-and-conquer video generation framework~\cite{videoldm,nuwa,long-survey,long-transformer,beyond-clip}, by first computing low FPS anchor frames in the Coarse process, and completing the video with high FPS video with interpolation models. However, artifacts and uncertainty in the predicted anchor frames that are caused by loss of temporal coherence, remain a hindrance of fidelity. To this end, we introduce a Coarse-to-(Re)fine module, which kills two birds with one stone. This module is capable of rectifying artifacts of the predicted anchor frames predicted in the Coarse process, meanwhile, enhancing the overall smoothness of the long video in the (Re)fine generation process. Without loss of generality, (Re) stands for the correction of anchor frames and fine process for the generation of high FPS frames. Furthermore, we develop a Temporal Progressive Denoising Scheduler module that enriches temporal information throughout the generation pipeline and improves the overall fidelity. In total, our contributions are three-fold:


\begin{itemize}
    \item We propose MiLA, a high-fidelity long-term video generation world model framework, which is tightly controlled by control signals and previous frames. Through our coarse-to-(Re)fine process, the framework successfully produces extended driving scenes lasting up to one minute without degradation of fidelity.
    \item To further explore the potential of our framework of long video generation, we incorporate a Joint Denoising and Correcting Flow (\textbf{JDC}) approach that addresses imperfections in predicted anchor frames while optimizing overall video smoothness. Furthermore, we propose a temporal Progressive Denoising Scheduler module (\textbf{TPD}) to enhance the fidelity of final frames with the guidance of previously generated frames within the batch.
    \item We validate our approach through comprehensive experiments, demonstrating state-of-the-art performance in both short and long-video generation on the nuScenes.  
\end{itemize}
\vspace{-2mm}

\section{Related Work}
\label{sec:related_work}
\vspace{-1mm}

\begin{figure*}[!htbt]
    \centering
    \includegraphics[width=1.0\linewidth]{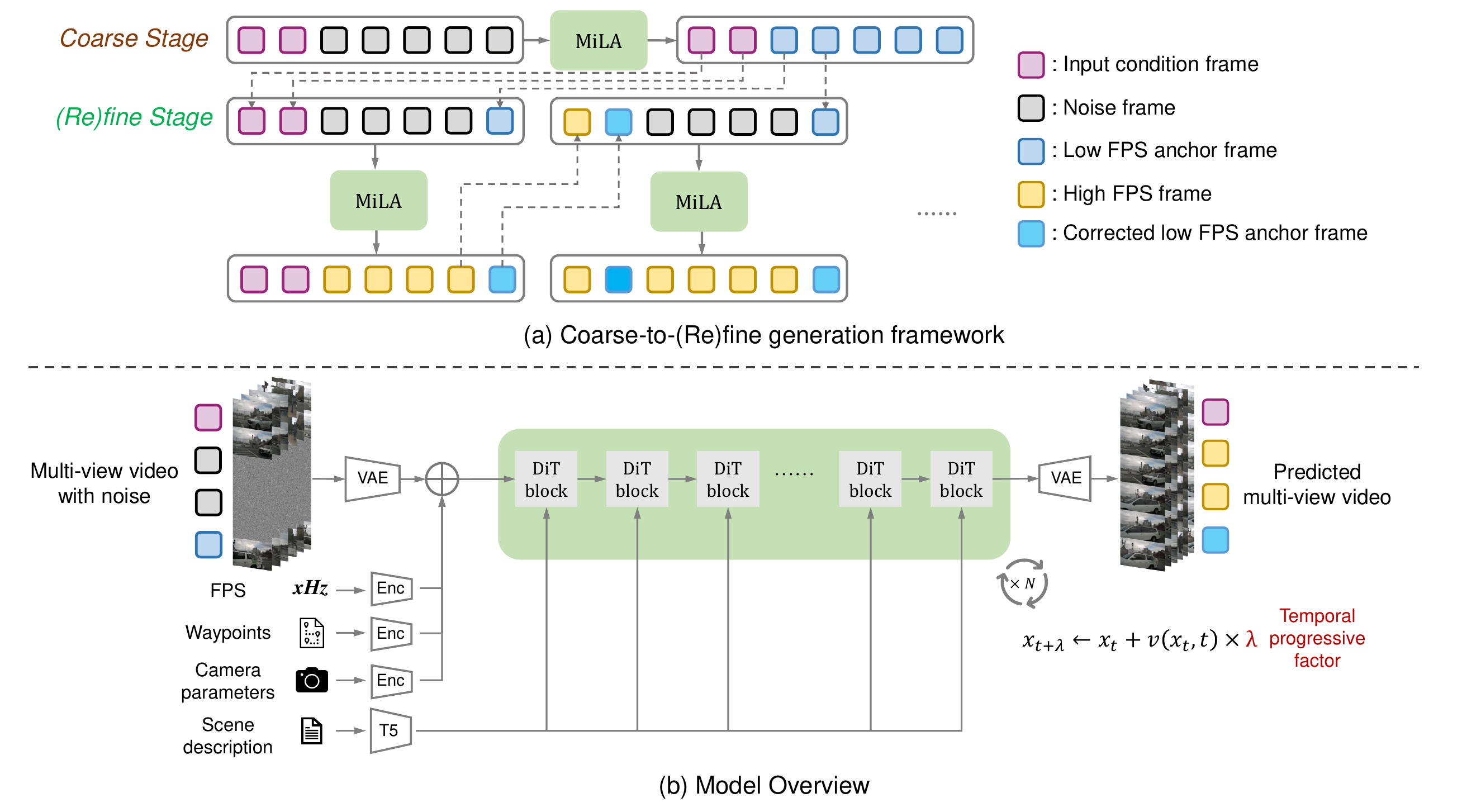}
    \vspace{-5mm}
    \caption{The overall pipeline of the proposed MiLA. \textbf{\textit{Top:}} The Coarse-to-(Re)fine generation framework. The conditions of the current (Re)fine stage involve both the corrected high FPS frames from the previous (Re)fine stage and the predicted low FPS anchor frames from the Coarse stage. The low FPS anchor frames are jointly denoised with high FPS interpolation frames to correct the artifacts. \textbf{\textit{Bottom}}: The generation model structure of MiLA. The inputs of the model include multi-view video (condition frames, noise, and anchor frames occasionally), FPS, waypoints, camera parameters, and brief textual scene descriptions. After encoding multi-modal inputs with different encoders, a DiT-based framework denoises the noised video tokens for $N$ times and outputs the prediction video.}
    \label{fig:framework}
    \vspace{-2mm}
\end{figure*}

\subsection{World Models}

World Models refer to generative models that can learn to simulate an environment and predict its dynamics, making it possible to train agents within these simulated spaces~\cite{world-model,genie,sketch2scene, wm-survey}. It is widely utilized for game simulation~\cite{genie,sketch2scene,dreamtocontrol,hafner2023mastering}, embody agent~\cite{embody-wm,openeqa,3d-vla,robodreamer,pivot-r,combo}, and Autonomous driving scenarios~\cite{drivedreamer,drivewm,adriver,vista,occworld,lidardm}. 
The world model in Autonomous driving can be generally separated into two paradigms according to the prediction format. The first type typically
predicts the future 3D representation of the driving scenes. Vidar~\cite{vidar} and LidarDM~\cite{lidardm} propose to estimate the future lidar points to forecast the geometric information. Several other works~\cite{cam4docc, occworld, driving_in_the_occupancy_world} employs future occupancy prediction as the basis of building a world model. Another line of work generates RGB videos in incoming seconds. Because of the scalability and data accessibility, such vision world model paradigm attracts more attention. DriveDreamer~\cite{drivedreamer} and its extended work~\cite{drivedreamer2} utilized HD-maps, bounding boxes, and reference frames of the initial timestamp and predicted the future states aggressively. ADriver-I~\cite{adriver} proposed a self-regressive generation strategy by leveraging the ability of the waypoints prediction module within the framework. Recently, Vista~\cite{vista} developed a high-resolution, long-term world model framework with notable fidelity enhancement modules, but it still struggles to generate undistorted long-term video.
\vspace{-1mm}


\subsection{Long-term Video Generation}
\vspace{-1mm}

Existing approaches for long-term video generation can be split into two distinct paradigms, as illustrated in Fig.~\ref{fig:comparison_RVD}. Specifically, most Autonomous driving world models follow the Recurrent generation paradigm~\cite{vista,wovogen,synthesizing_coherent,grimm-open-ended}, also known as the Auto-regressive framework. In this framework, each new batch is generated with the condition on the preceding batch by replacing the initial token with the condition token. To ensure this framework's quality, recent work~\cite{p-regressive} proposed an asynchronous diffusion process and divides the batch generation into frame generation. The first frame within a batch is first diffused and pops out consequently. At the same time, a new noise token is placed at the end of the batch to maintain the batch size. However, the accumulated error problem is not properly addressed.  

On the other hand, some researchers~\cite{cogvideo,videoldm,beyond-clip} are devoted to Divide-and-Conquer paradigm that generates anchor frames and interpolation frames with two distinct processes. Nuwa-Xl~\cite{nuwa} further improved the model structure for better performance. To further exploit the potential, studies~\cite{generating_long} leverage this paradigm for resolution enhancement tasks. Recent work~\cite{Adobe_temporal} explores asynchronous temporal sampling strategies for video generation, though consistency inner clip remains unaddressed. Critically, however, limited attention has been paid to artifacts inherent in predicted anchor frames, leaving a gap in understanding their impact on overall video quality.

\begin{figure*}[ht]
\centering
\begin{subfigure}
{.66\linewidth}
  \centering
  \includegraphics[width=\linewidth]{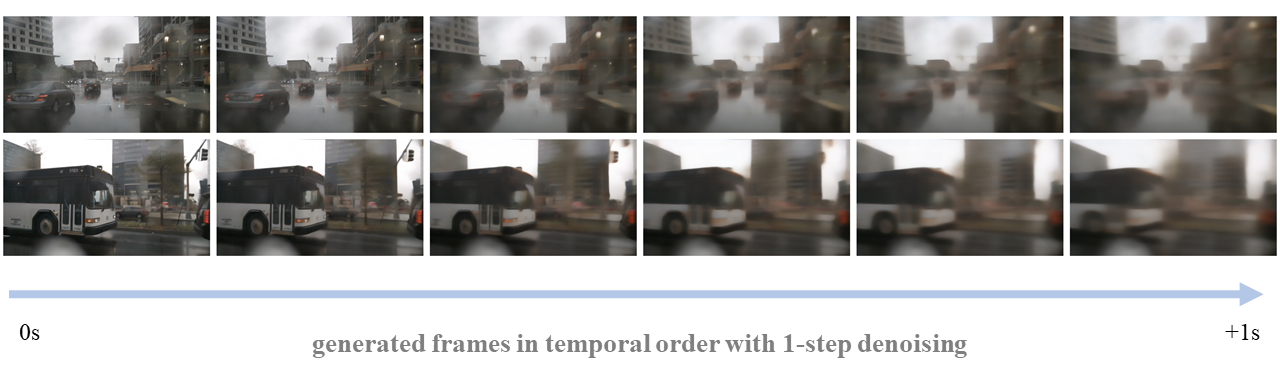}
  \caption{Samples of one-step denoising.}
  \label{fig:one_step_vis}
\end{subfigure}%
\begin{subfigure}{.24\linewidth}
  \centering
  \includegraphics[width=\linewidth]{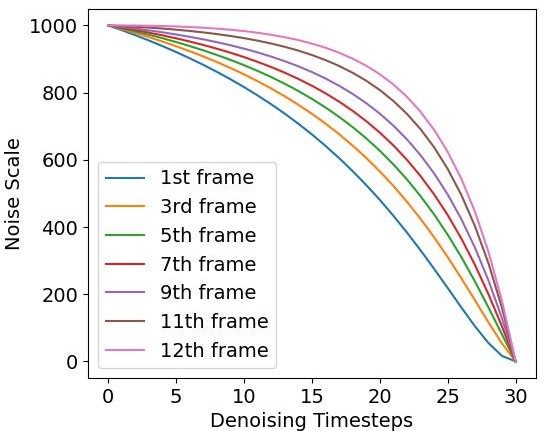}
  \caption{Noise scales of TPD}
  \label{fig:noise_steps}
\end{subfigure}
\caption{\textbf{\textit{(a): }}The predicted frames of two distinct views were generated in increasing temporal order, where frames of the Front View are placed in the first row and frames of the Back Left View are placed in the second row, respectively. All frames are produced using a single diffusion step. Frames that are closer to the condition frames are less degradative. \textbf{\textit{(b): }}A sample visualization of the function used to compute the noise scale, based on the noisy frame index and denoising timestamps as variables.}
\label{fig:noise_steps_all}
\end{figure*}
\begin{figure*}[ht]
    \centering
    \includegraphics[width=1.0\linewidth]{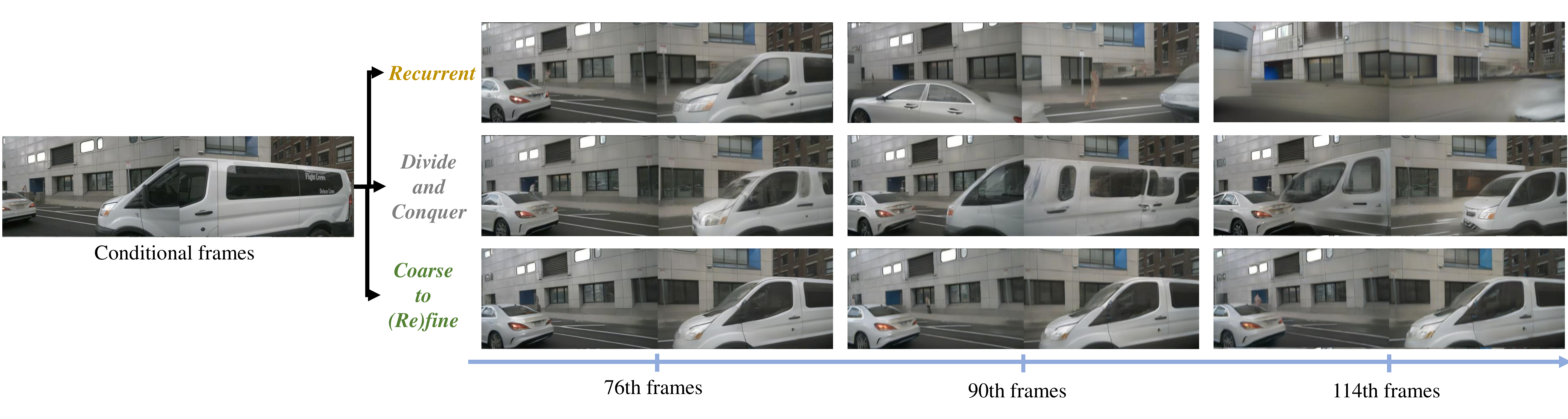}
    \caption{Comparison between Recurrent~\cite{videoldm,nuwa} Divide-to-Conquer~\cite{vista,Gaia-1} and the proposed Coarse-to-(Re)fine generation frameworks with two distinct views (front-right and back-right) and three frames (76th, 90th and 114th). All frames are generated with our MiLA and different paradigms for long video generation are adopted. \textbf{\textit{Left: }}The condition frames for video generation. The predicted videos are placed to \textbf{\textit{Top Right: }}Recurrent, \textbf{\textit{Middle Right: }}Divide-and-Conquer and \textbf{\textit{Bottom Right: }}Coarse-to-(Re)fine. }
    \label{fig:comparision_refine}
    \vspace{-2mm}
\end{figure*}

\section{Method}

Fig.~\ref{fig:framework} introduces the overview of our proposed MiLA, provides an overview of our proposed MiLA framework, which enables the generation of long-term videos based on user-defined reference frames and inputs, including waypoints, camera parameters, temporal frame indices, and scene descriptions. The structure of this section is as follows. We begin by briefly introducing the design of our model in Sec.~\ref{sec:model} and the overall framework for long-video generation in Sec.~\ref{sec:long}, respectively. We next introduce the correction module (JDC) of the (Re)fine process. This component, which integrates error rectification and noise reduction into a unified framework, is described in detail in Sec.~\ref{sec:correct}. Additionally we conclude the temporal enhancement  design (TPD) in Sec.~\ref{sec:scheduler} and loss function in Sec.~\ref{sec:loss}, respectively.


\subsection{Model Overview}
\label{sec:model}
\vspace{-1mm}

We adopt the framework of Open-Sora~\cite{opensora}, a DiT-based video generation model, as the backbone architecture of MiLA. Our MiLA predicts $x_{0:V-1}^{N:N+S-1}$ multi-view frames conditioned on $x_{0:V-1}^{0:N-1}$, where V, N, S represent the number of views, conditional frames and total noisy frames, respectively, as illustrated in Fig.~\ref{fig:framework}. To simplify the expression, we omit the use of the $0:V-1$ notation for multi-view representation throughout the remainder of the paper. 
\vspace{-3mm}

\paragraph{Flexible Condition Embedding.} To guide generation with waypoints and camera view guidance, we use Fourier embedding followed by Multiple-Layer-Perception (MLP) layers to encode the input conditional parameters into embeddings. Formally, we fuse all camera parameters and waypoints $c=\{c^0, c^1, ..., c^{N+S-1}\}$ with frames embedding $h_f$ as follow, 
\begin{align}
    &h_c=[h_c^0, h_c^1, ..., h_c^{N+S-1}]=\text{MLP}(\text{FFT}(c)),\\
    &h_{fuse}=h_{f}+h_c.
\end{align}

\paragraph{Multi-view Enhanced Spatial Attention.} 

We achieve multi-view well-aligned generation by expanding the spatial receptive field of DiT blocks to include all views' patches.
\begin{equation}
    h_{spatial} (Q,K,V) = \text{softmax}(\frac{QK^T}{\sqrt{d}})\cdot V,
\end{equation}
where $d$ denotes the dimension of embeddings, and $Q, K, V\in \mathbb{R}^{T\times (VHW)\times C}$ denotes feature maps reshaped and projected from the input frame embeddings $h_{frame}\in \mathbb{R}^{T\times V\times H\times W\times C}$.
\vspace{-2mm}
\paragraph{Preliminary: Rectified Flow.}
Rectified flow~\cite{rflow} is an ODE-based approach for transferring between two distributions $x_0\sim \pi_0$ and $x_1\sim \pi_1$.  In rectified flow, an interpolated middle-stage result can be represented as
\begin{equation}
    \label{eqa:rflow-base}
    x_t = (1-t)x_0 +  t x_1, 
\end{equation}
where $t \in [0,1]$ is the timestamp of distribution transferring. 
Specifically, in diffusion models, $x_0$ is typically a standard gaussian distribution, namely $x_0\sim \mathcal{N}(0,1)$. 
Therefore, we can transform the Equation. \ref{eqa:rflow-base} into
\begin{equation}
    x_t = (1-\alpha)x_1 +  \alpha x_0, x_0\sim \mathcal{N}(0,1),
\end{equation}
where $\alpha=1-t$ is the noise intensity. Without loss of generality, a straight trajectory results in fast inference, and 
\begin{equation}
    \frac{\mathrm{d}}{\mathrm{d}t}x_t = v(x_t, t).
\end{equation}
With the above equation, we can predict the target $\hat{x}_1$ in the following way:
\begin{equation}
    \hat{x}_1 = x_0+\int_0^1 v(x_t, t) \mathrm{d}t.
\end{equation}
\vspace{-2mm}
\subsection{Long-term Video Generation Framework}
\label{sec:long}



An illustration of our Coarse-to-(Re)fine pipeline is placed in Fig.~\ref{fig:framework}. Our Coarse process adopts a divide-and-conquer approach by first predicting anchor frames as references, using low frame-rate embeddings. Instead of interpolating from a single start and tail frame, our (Re)fine process employs a recurrent strategy, as shown in Fig.~\ref{fig:framework}. More specifically, we recurrently leverage multiple conditions frames, incorporating both the high FPS frames from the previous (Re)fine process and the low FPS anchor frames from the Coarse process. This approach enables our model to better capture motion dynamics while simultaneously understanding the broader environment.

Additionally, we incorporate a correction module within the (Re)fine process to adjust any unrealistic motion in the anchor frames. A unique noise-adding and denoising approach is applied to enhance the smooth transitions between anchor frames. Further design details and analysis are provided in the following Sec.~\ref{sec:correct}.   

\subsection{Joint Denoising and Correcting Flow}
\label{sec:correct}

The Divide-and-conquer framework achieves long-term video generation fidelity through its reliance on anchor frames, yet this strength introduces a critical trade-off: diminished temporal coherence between frames, as illustrated in Fig.~\ref{fig:comparision_refine}. While anchor frames preserve structural consistency over extended sequences, their sparse temporal distribution weakens inter-frame dependencies, often manifesting as artifacts such as unstable object motion and reduced video smoothness. To address this inherent tension between fidelity and fluidity, we propose the Joint Denoising and Correcting Flow, a module designed to harness the advantages of anchor frames while actively mitigating their limitations. This approach synchronously optimizes low-FPS anchor frames (to enhance structural fidelity) and high-FPS interpolated frames (to restore temporal continuity), thereby resolving motion inconsistencies.

We begin by establishing a theoretical link to the principles of latent diffusion models (LDM), positing that the noise inherent to anchor frames can be decomposed into structured and stochastic components
\begin{equation}
    z=(1-\alpha_1)x_0+\alpha_1n_1, n_1\sim \mathcal{N}(0,1),
    \label{eqa:anchor}
\end{equation}
where $z$ and $x_0$ represented the predicted and ground truth anchor frame, respectively. $\alpha_1$ indicates the noise scale. Given the above assumption, we wish to develop a noisy predicted anchor frame with a mixture of the predicted anchor frame and Gaussian noise. 
\begin{equation}
    x_n=(1-\alpha_2)z+\alpha_2n_2,n_2\sim \mathcal{N}(0,\sigma^2),
    \label{eqa:noise_anchor}
\end{equation}
where $\alpha_2$ and $\sigma$ represent the noise scale and standard deviation, which are the key values of the noise-adding procedure. By inserting Equation.~\ref{eqa:anchor} into \ref{eqa:noise_anchor}, we can have
\begin{equation}
    x_n=(1-\alpha_2)(1-\alpha_1)x+\alpha_2n_2+\alpha_1(1-\alpha_2)n_1.
\end{equation}
Drawn from the standard LDM, the noise added to the noisy predicted anchor frame should be a strand normal distribution. To this end, we compute the standard deviation $\sigma$ as
\begin{equation}
    \label{eqa:sigma}
    \sigma^2=\frac{\alpha_2-2\alpha_1\alpha_2+2\alpha_1}{\alpha_2}.
\end{equation}
Please refer to supplementary material for detailed derivation. In other words, the noise we add to the predicted anchor frames satisfies that $n_2\sim \mathcal{N}(0, \frac{\alpha_2-2\alpha_1\alpha_2+2\alpha_1}{\alpha_2})$.


\begin{table*}[h]
    \centering
    \begin{tabular}{c|c|cccc}
        \hline
         Method & Num Samples & $\text{FID}_{front}\downarrow$ & $\text{FVD}_{front}\downarrow$ & $\text{FID}_{multi}\downarrow$ & $\text{FVD}_{multi}\downarrow$ \\
         \hline
         DriveDreamer~\cite{drivedreamer} & - & 52.6 & 452.0 & - & - \\
         ADriver-I~\cite{adriver} & - & \textcolor{blue}{5.5} & 97.0 & - & - \\
         WoVoGen~\cite{wovogen} & -  & - & - & 27.6 & 417.7 \\
         DriveWM~\cite{drivewm} & - & - & - & 15.8 & 122.7 \\
         GenAD~\cite{genad} & - & 15.4 & 184.0 & - & - \\
         DriveDreamer-2~\cite{drivedreamer2} & - & - & - & 11.2 & 55.7 \\
         Vista~\cite{vista} & 5369 & 6.9 & 89.4 & - & - \\
         \hline
         \hline
         
         MiLA (4f cond) & 600 & 8.9 & 89.3 & 4.9 & 36.3 \\
         MiLA (1f cond) & 5369 & \textcolor{red}{4.9} & \textcolor{blue}{42.9} & \textcolor{red}{3.0} & \textcolor{blue}{18.2} \\
         MiLA (4f cond) & 5369 & 5.9 & \textcolor{red}{36.8} & \textcolor{blue}{4.1} & \textcolor{red}{14.9} \\
         \hline
    \end{tabular}
    \caption{Comparison of generation fidelity on nuScenes validation set, where \textcolor{red}{red} and \textcolor{blue}{blue} represent the best and the second best values. MiLA outperforms the state-of-the-art world models with respect to generation quality.}
    \vspace{-4mm}
    \label{tab:sota}
\end{table*}

\begin{table*}[h]
    \centering
    \begin{tabular}{c|c|cccc}
        \hline
         Method & Num Samples & $\text{FID}_{0-10s}\downarrow$ & $\text{FVD}_{0-10s}\downarrow$ & $\text{FID}_{8-10s}\downarrow$ & $\text{FVD}_{8-10s}\downarrow$ \\
         \hline
         Vista~\cite{vista} & 150 & 31.9 & 666.2 & 71.0 & 679.3 \\
         
         MiLA & 150 & \textcolor{red}{18.3} & \textcolor{red}{331.9} & \textcolor{red}{42.6} & \textcolor{red}{358.7} \\
         \hline
    \end{tabular}
    \caption{Comparison of generation fidelity on nuScenes validation set, where \textcolor{red}{red} represent the best value. MiLA outperforms the state-of-the-art world models with respect to generation quality.}
    \vspace{-4mm}
    \label{tab:sota_long}
\end{table*}
To this end, we denote the function for deducing the noise timestamp of the anchor frames given an origin timestamp $t$ as $g(t)$, and the $\alpha_2$ can be calculated as
\begin{equation}
    \begin{aligned}
        g(t) &=1-(1-\alpha_1)(1-\alpha_2),\\
        \alpha_2&=\frac{g(t)-\alpha_1}{1-\alpha_1}.
    \end{aligned}
\end{equation}


To simplify the overall noise-adding process, we define noise scale $\alpha_1$ to be constant and $g(t)$ to be a linear function of $\alpha_0$, and hence $\alpha_2$ can be computed. Next, according to Equation.~\ref{eqa:sigma}, the standard deviation $\sigma$ presented in Equation.~\ref{eqa:noise_anchor} can be derived. Finally, the noise-adding process of the predicted anchor frames can be determined with noise scale $\alpha_2$ and noise variance $\sigma^2$.   




\subsection{Temporal Progressive Denoising Scheduler}
\label{sec:scheduler}






\begin{figure*}[!t]
    \centering
    \includegraphics[width=\linewidth]{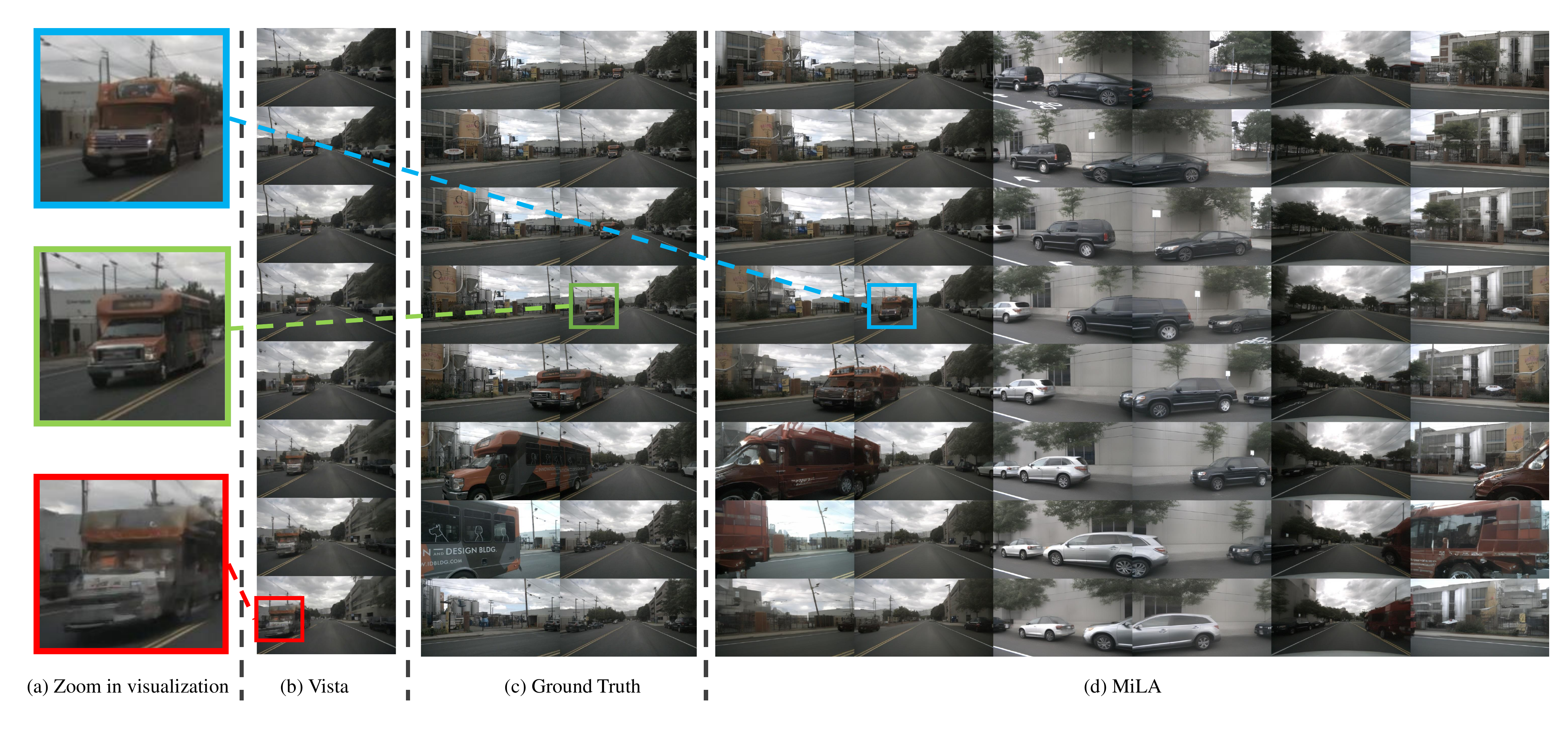}
    \vspace{-8mm}
    \caption{Multiple view generation. Columns from left to right record the frames of \textbf{\textit{(a: )}} Zoom in visualization, \textbf{\textit{(b: )}} Vista, \textbf{\textit{(c: )}} Ground Truth of Front Left and Front view  and \textbf{\textit{(d: )}} our MiLA, respectively.}
    \label{fig:multi_view}
    \vspace{-2mm}
\end{figure*}

To enhance the fidelity of the overall video, we design a specific denoising scheduler. Our key insight is demonstrated in Fig. \ref{fig:one_step_vis}. The frames closer to the condition frames tend to have better fidelity with a single denoising step. This phenomenon raises an interesting assumption, is it possible to predict earlier frames with fewer steps? Based on this assumption, we can easily suggest another hypothesis. Whether the quicker revelation of preceding frames provides more detailed information for following frames and hence enhances the generation quality of later frames. 

To properly utilize these observations, we design a function with a noisy temporal frame index of the collection of all noisy frames $s \in S$ and denoising timestamps $t$ as variables and compute the denoise scheduler coefficient. The function should meet the requirements:
\begin{itemize}
    \item All frames within one denoising batch must reach the final denoised status simultaneously. 
    \item Earlier frames are denoised with a larger scale at the first few steps and result in an almost-denoised status promptly.
    \item The denoise scheduler of trailing frames should follow a slow-fast pattern, which dramatically improves in scale in the last few steps. 
\end{itemize}
To this end, we derive a function to transform the original timestamp $t$ in the following way:
\begin{equation}
    f(t,\phi_s) = 1- \frac{\cos{((1-t)\omega+\phi_s)} - \cos{(\phi_s)}}{\cos{(\omega+\phi_s)} - \cos{(\phi_s)}},
\end{equation}
where $\phi_s$ is computed from the $s$ with normalization. 
\begin{equation}
    \phi_s = 0.5\pi \times \frac{s}{|S|},
\end{equation}
and the visualization of the function is shown in Fig.~\ref{fig:noise_steps}. The computation of the scale coefficient is adapted from a simple $\cos$ function with all values among the function constrained with $0$ and $\pi$. A mono decreasing properly is utilized and various initialization between $0$ and $\pi$ would result in various decreasing trends. Then a normalization term is employed to constrain the trends.

\subsection{Loss Function}
\label{sec:loss}
During the training stage, we randomly sample one timestamp for each iteration, and the overall minimization target of MiLA over the predicted flow $v$ is as follows, 
\begin{equation}
\begin{split}
    \
    \sum_s\int_0^1 &\mathbb{E}[(1-m_s)\odot \Vert y - v(x_{f(t, \phi_s)}, f(t, \phi_s)) \Vert^2
    \\
    &+ m_s\odot \Vert y - v(x_{g(t)}, g(t) \Vert^2]\mathrm
    {d}t,
\end{split}
\end{equation}
where $y=(x_1-x_0)$ denoted the flow vector between starting noise and ground truth frames' latent embeddings, $v(x_t, t)$ is the flow predicted by our model, $m_s$ is the binary mask indicating noisy anchor frames. 



\begin{figure}[!t]
    \centering
    \centering
    \begin{subfigure}
    {.5\linewidth}
  \centering
  \includegraphics[width=\linewidth]{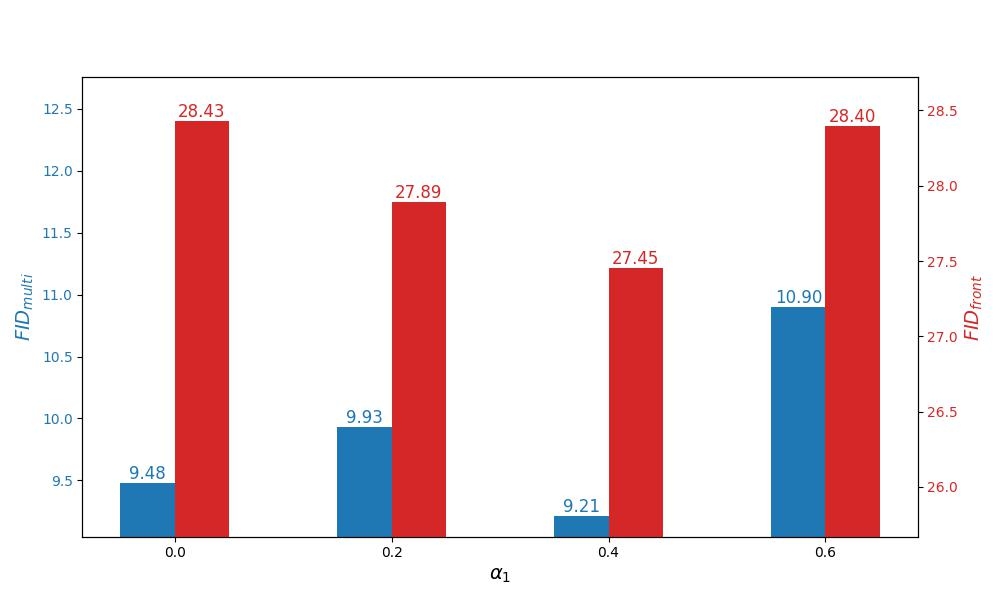}
  \caption{Ablation studies of $\alpha_1$}
  \label{fig:one_step_vis}
\end{subfigure}
\hfill
\begin{subfigure}
    {.5\linewidth}
  \centering
  \includegraphics[width=\linewidth]{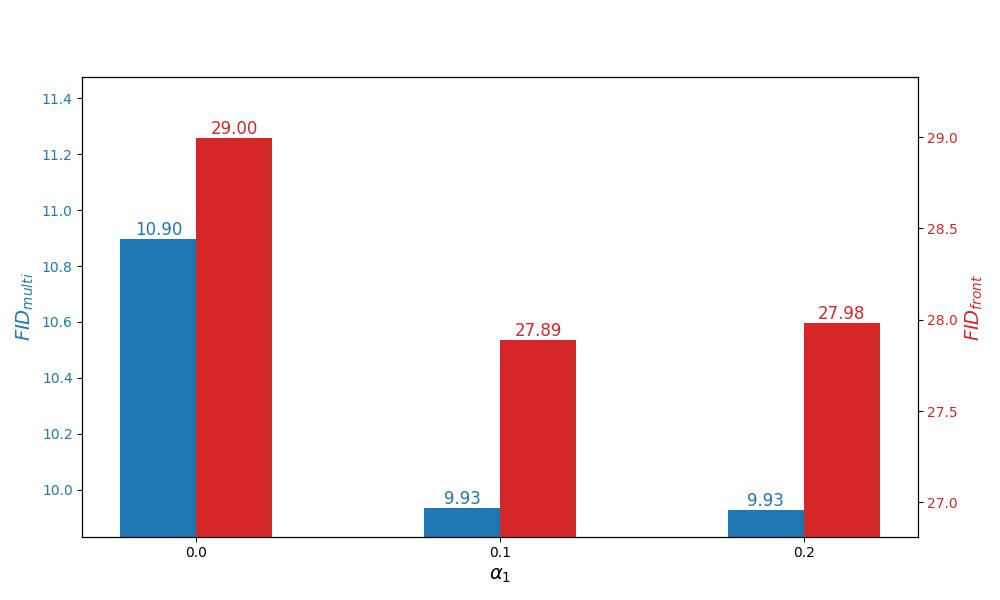}
  \caption{Ablation studies of $\alpha_2$}
  \label{fig:one_step_vis}
\end{subfigure}

    \vspace{-2mm}
    \caption{Ablation of $\alpha$, where \textcolor{blue}{blue} and \textcolor{red}{red} denotes \textcolor{blue}{$\text{FID}_{multi}$} and \textcolor{red}{$\text{FID}_{front}$}, respectively}
    \label{fig:alpha_abla}
    \vspace{-2mm}
\end{figure}
\section{Experiments}
\label{sec:experiments}
In this section, we start by describing the experimental setup with implementation details, training \& inference pipeline, dataset, and evaluation metric in Sec.~\ref{sec:exp_setup}. Detailed analysis of both the qualitative and quantitative results are then conducted in Sec.~\ref{sec:result}. Finally, we will conduct several ablation studies and analyses on our key modules in Sec.~\ref{sec:ablation}. 
\subsection{Experimental Setup}
\label{sec:exp_setup}
\begin{table*}[t]
    \centering
    \begin{tabular}{c|c|c|cccc}
        \hline
         Progressive & Num & \multirow{2}{*}{Generation Mode} & \multirow{2}{*}{$\text{FID}_{front}\downarrow$} & \multirow{2}{*}{$\text{FVD}_{front}\downarrow$} & \multirow{2}{*}{$\text{FID}_{multi}\downarrow$} & \multirow{2}{*}{$\text{FVD}_{multi}\downarrow$} \\
         Denoising & Samples &  & & & & \\
         \hline
         $\times$ & 600 & $4f\times12fps\rightarrow 12f\times 12fps$ & 9.1 & 81.3 & 4.6 & 32.2 \\
         $\checkmark$ & 600 & $4f\times12fps\rightarrow 12f\times 12fps$ & \textcolor{red}{8.0} & \textcolor{red}{63.6} & \textcolor{red}{3.6} & \textcolor{red}{23.8} \\
         \hline
         $\times$ & 600 & $4f\times12fps\rightarrow 24f\times 12fps$ & 9.5 & 106.4 & 6.3 & 45.6 \\
         $\checkmark$ & 600 & $4f\times12fps\rightarrow 24f\times 12fps$ & \textcolor{red}{8.9} & \textcolor{red}{89.3} & \textcolor{red}{4.9} & \textcolor{red}{36.3} \\
         \hline
         \hline
         $\times$ & 150 & $4f\times12fps\rightarrow 12f\times 1fps$ & 24.9 & 317.1 & 10.1 & 81.8 \\
         $\checkmark$ & 150 & $4f\times12fps\rightarrow 12f\times 1fps$ & \textcolor{red}{23.0} & \textcolor{red}{271.8} & \textcolor{red}{9.4} & \textcolor{red}{73.5} \\
         \hline
    \end{tabular}
    \caption{Impacts of our Temporal Progressive Denoising Scheduler, where \textcolor{red}{red} represent the best value. Our scheduler improves generation quality in various durations. Videos with 12 frames are generated based on 4 condition frames. The previous process is repeated to generate another 12 frames, based on the last 4 frames for a 24 frames video.}
    \label{tab:ablation_study}
\end{table*}
\begin{table*}[t]
    \centering
    \begin{tabular}{ccc|cccc}
        \hline
         TPD & Anchor Frame & JDC & $\text{FID}_{front}\downarrow$ & $\text{FVD}_{front}\downarrow$ & $\text{FID}_{multiview}\downarrow$ & $\text{FVD}_{multiview}\downarrow$ \\
         \hline
           & & & 9.1 & 81.3 & 4.6 & 32.2 \\
           $\checkmark$ & & & 8.0 & 63.6 & 3.6 & 23.8 \\
           $\checkmark$ & GT & & 5.5 & 38.1 & 2.4 & 11.8 \\
           $\checkmark$ & pseudo & & 8.9 & 85.7 & 3.9 & 28.5 \\
           $\checkmark$ & pseudo & $\checkmark$ & 7.7 & 60.4 & 3.3 & 18.8 \\
         \hline
    \end{tabular}
    \caption{Ablations of the components in MiLA short-term generation.}
    \label{tab:ablation_big}
    \vspace{-3mm}
\end{table*}
\paragraph{Dataset and Evaluation Metrics.}
We conduct all experiments on the nuScenes~\cite{nuscenes} dataset, which includes 700 training and 150 validation scenes collected in Boston and Singapore, with each scene spanning around 20 seconds and containing high-definition 360-degree images. Additionally, Fréchet Inception Distance (FID)~\cite{fid} and Fréchet Video Distance (FVD)~\cite{fvd} are utilized for evaluating the overall generation quality. Specifically, A lower FID and FVD score indicate the closer similarity of image and video, respectively. To comprehensively evaluate various approaches that produce videos with different views, we extend FID and FVD to the following metrics. Where $\text{FID}_{front}$ and $\text{FVD}_{front}$ are used to assess the generation quality of front-view videos, and $\text{FID}_{t_0-t_1s}$ and $\text{FVD}_{t_0-t_1s}$ are used to assess the front-view generation quality from $t_0$ to $t_1$ second. Moreover, for computing FID and FVD, we employ a pre-trained Inception-v3 network~\cite{inception}, which encodes images into 2048-dim features, and I3D network~\cite{i3d} for feature extraction respectively. 
\vspace{-2mm}
\paragraph{Training and Evaluation Details.}
More detail information regarding the training and inference procedures is provided in the supplementary material.

\subsection{Evaluation Results}
\label{sec:result}

\vspace{-1mm}

\paragraph{Qualitative Results.}
Fig.~\ref{fig:multi_view} demonstrates the overall effectiveness of MiLA in generating high-fidelity, multi-view driving scene videos. Experiments prove our MiLA is capable of producing spatially and temporally consistent video across different cameras and time indices. More specifically, the outline predicted by our MiLA significantly outperforms Vista as shown in the Zoom-in patch. We further demonstrate the overall quality of long-term video generated by our MiLA in comparison with Recurrent and Divide-and-Conquer frameworks, as in Fig.~\ref{fig:comparision_refine}. Compared to the recurrent generation framework, our Coarse-to-(Re)fine framework maintains overall fidelity without noticeable degradation, especially in long-term video generation. Furthermore, our design enhances structural detail for dynamic objects, effectively reducing distortions of the white van. Please refer to our supplementary material for more comprehensive qualitative and quantitative results and studies of consistency and object kinematics.

\paragraph{Quantitative Results.} To fully assess the ability of MiLA, we conduct experiments across various methods for both front and multi-view generation. 
We add waypoints, camera parameters, and brief text descriptions as conditions for multi-view video generation. 
To make the greatest effort possible for fair comparison, we generate 28 frames at 12 FPS conditioned on input 4 or 1 condition frames with the first 25th frames are used for evaluation. However, methods in this field employ diverse settings in terms of duration, dataset length, and conditional inputs, making it challenging to conduct a fair and comprehensive comparison across all methods, the single-frame conditioning experiment is totally aligned with Vista’s setting.  

Comprehensive quantitative metrics of all compared approaches on nuScenes validation set are recorded in Table~\ref{tab:sota}. We all significantly surpass the current state-of-the-art approaches in both FID and FVD in both front-view and multiple-views. More specifically, we surpass DriveDreamer-2 by $7.1$ in FID and $40.8$ in FVD for multi-view generation. On the other hand, our MiLA surpasses Vista by $2.0$ in FID and $46.5$ in FVD in a totally fair comparison. Additionally, Table~\ref{tab:sota_long} recorded the metric for long video generation, where the proposed MiLA significantly outperform current approaches. Please refer to our supplementary material for more quantitative results.



\subsection{Ablation Studies}
\label{sec:ablation}
\paragraph{Effect of design module.}
To thoroughly evaluate the proposed MiLA module, we conduct comprehensive ablation studies, as summarized in Table~\ref{tab:ablation_big}. Here, TPD, and JDC denote the inclusion of these modules in the framework. In the Anchor Frame column, blank, GT (ground truth), and pseudo correspond to scenarios involving low-FPS anchor frames: blank indicates the absence of anchor frames, GT uses ground truth frames, and pseudo employs predicted frames. Obviously, our experiments demonstrate that integrating TPD and JDC effectively improves overall generation quality,
\vspace{-3mm}
\paragraph{Effect of JDC.} As illustrated in Fig.~\ref{fig:alpha_abla}, the parameter $\alpha_1$ and $\alpha_2$ critically governs the correction efficacy of anchor frames. Specifically, excessive noise amplitude compromises scene structural integrity, while insufficient noise fails to account for dynamic object deformation. This trade-off necessitates careful calibration of noise level to balance artifact suppression and motion fidelity, as validated by our quantitative and qualitative analyses. 
\vspace{-3mm}
\paragraph{Effect of TPD.}
Table~\ref{tab:ablation_study} record the effect of the proposed TPD. Specifically, we perform comparisons on generating videos with varying frame numbers and frame rates, utilizing the same FID and FVD metrics employed in previous experiments. The results consistently highlight the effectiveness of the designed progressive denoising Scheduler.

\section{Conclusion}

In this paper, we introduce MiLA, a video generation world model framework that generates driving scenes' videos given prior frames and future waypoints. To achieve this, we propose a coarse-to-(Re)fine framework to generate low-FPS anchor frames first, then interpolate high-FPS frames and restore anchor frames jointly. We design a JDC and a TPD module to enhance temporal consistency. Lastly, we hope MiLA enlightens practitioners concerning long-term high-fidelity world models.


\small
    \bibliographystyle{ieeenat_fullname}
    \bibliography{main}

\clearpage
\setcounter{page}{1}
\maketitlesupplementary
\setcounter{section}{0}
\renewcommand\thesection{\Alph{section}}



\section{Detail Supplementary}

\subsection{Equation Derivation}
Here we present the detailed derivation of the noise-adding procedure within the predicted low FPS anchor frames. The main challenge is the computation of $\alpha_2$ and $\sigma^2$ in Equation. \textcolor{cyan}{11} in the main paper. We start with Equation .\textcolor{cyan}{12} in the main paper, 
\begin{equation}
n_3 = \alpha_2n_2+\alpha_1(1-\alpha_2)n_1
\end{equation}
where
\begin{equation}
\label{eqa:n3_1}
    n_3 \sim \mathcal{N}(0,(\alpha_2)^2\sigma^2 + (\alpha_1)^2(1-\alpha_2)^2).
\end{equation}
To strictly follow the transferring process of rectified flow~\cite{rflow} in Equation. \textcolor{cyan}{5} in the main paper, where the noise part is strictly constrained to be a standard Gaussian distribution $n \sim \mathcal{N}(0,1)$, we can rewrite the noise part of Equation. \textcolor{cyan}{12} in the main paper to be
\begin{equation}
    n_3 = (1-(1-\alpha_2)(1-\alpha_1))n.
\end{equation}
We can rewrite the distribution of $n_3$ as 
\begin{equation}
\label{eqa:n3_2}
        n_3 \sim \mathcal{N}(0,(1-(1-\alpha_2)(1-\alpha_1))^2).
\end{equation}
Then, we can simply derive the following equation by combining Equation.~\ref{eqa:n3_1} and~\ref{eqa:n3_2} in the supplementary.
\vspace{-3mm}

\begin{equation}
    (\alpha_2)^2\sigma^2 + (\alpha_1)^2(1-\alpha_2)^2 = 1-(1-\alpha_2)(1-\alpha_1))^2.
\end{equation}
The $\sigma^2$ can be computed
\begin{equation}
\begin{aligned}
    \sigma^2 &= \dfrac{1-(1-\alpha_2)(1-\alpha_1))^2 - (\alpha_1)^2(1-\alpha_2)^2}{(\alpha_2)^2} \\
             &=\frac{\alpha_2-2\alpha_1\alpha_2+2\alpha_1}{\alpha_2}
\end{aligned}
\end{equation}

\section{Implementation Detail}
\paragraph{Training.}
Videos with $16\times N (N\in [1,2,3])$ frames are inputs of MiLA's inputs, with the initial 1 or 4 frames for visual condition (as well as the last frame in serveral cases) and the remaining frames for prediction. 
\begin{table*}[t]
    \centering
    \begin{tabular}{c|c|c|cccc}
        \hline
         Trajectory & Num Samples & Num frames & $\text{FID}_{front}\downarrow$ & $\text{FVD}_{front}\downarrow$ & $\text{FID}_{multiview}\downarrow$ & $\text{FVD}_{multiview}\downarrow$ \\
         \hline
         $\times$ & 600 & 12 & 8.1 & \textcolor{red}{63.4} & 3.7 & 24.9 \\
         $\checkmark$ & 600 & 12 & \textcolor{red}{8.0} & 63.6 & \textcolor{red}{3.6} & \textcolor{red}{23.8} \\
         \hline
         $\times$ & 600 & 24 & \textcolor{red}{8.7} & 101.5 & 5.1 & 39.7 \\
         $\checkmark$ & 600 & 24 & 8.9 & \textcolor{red}{89.3} & \textcolor{red}{4.9} & \textcolor{red}{36.3} \\
         \hline
    \end{tabular}
    \caption{Impacts of trajectories' guidance, where \textcolor{red}{red} represent the best value. }
    \label{tab:ablation_study_traj}
    \vspace{-3mm}
\end{table*}
\begin{figure}[ht]
    \centering
    \includegraphics[width=\linewidth]{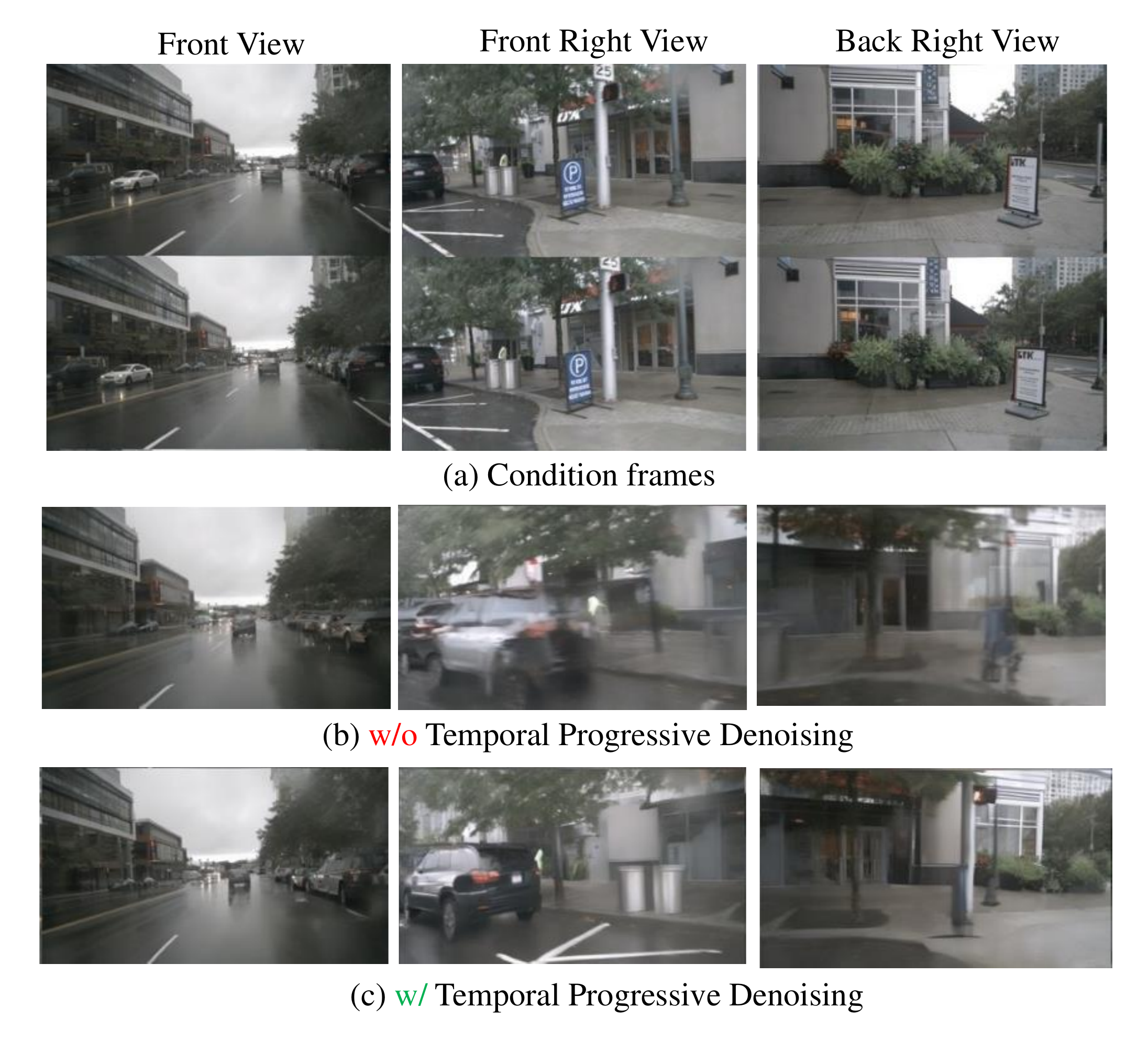}
    \caption{Effective of the usage of Temporal Progressive Denoising module over various views. The 12th predicted frames are presented in this figure. \textbf{\textit{Top:}} Condition frames used for video prediction. \textbf{\textit{Middle: }} Qualitative results without the usage of the Temporal Progressive Denoising module. \textbf{\textit{Bottom: }} Qualitative results with the usage of the Temporal Progressive Denoising module. The Temporal Progressive Denoising module effectively enhances clearness and consistency.}
    \label{fig:noise_scale_comp}
    \vspace{-5mm}
\end{figure}

All images are resized to $360\times 640$ to train MiLA, which has roughly 700M parameters in total (excluding VAE and text encoder T5~\cite{t5}). We load the pretained weight of Open-Sora 1.1~\cite{opensora}. Specifically, MiLA is trained in 3 stages on 64 NVIDIA A100 GPUs with a total batch size of 64. 
\paragraph{Fisrt Stage} In the first stage, we finetuned the model to predict the next 12 frames with consistent FPS within the clip (FPS randomly sampled from \{1,2,12\}). In this stage, we freeze most parameters and only train newly added modules (Multi-view Enhanced Spatial Attention module, Flexible Condition Embedding module), as well as the last DiT block as well as the output layer. This stages takes roughly 2k training steps.
\vspace{-3mm}
\paragraph{Second Stage} In the second stage, we fix the condition frames' FPS to 12 and train short-term as well as long-term MiLA with different generated frames' FPS. respectively. Most parameters, excluding a few embedding layers, are trainable. Both short-term and long-term model are trained for around 20k iterations in the second stage.
\vspace{-3mm}
\paragraph{Third Stage} In the third stage, we train the same parameters as the prior stage, with additional anchor frame correction training. To be specific, we set the tail anchor frame to noisy anchor frame and train the model to correct artifects with a rate of 75\%. MiLA is trained for about 8k steps in this stage.

In all stages, we set the learning rate to $1\times 10^{-4}$ with a linear warm-up of 1000 iterations. Textual descriptions and waypoints are dropped independently with a ratio of $15\%$ for classifier-free guidance~\cite{classifier-free-guidance} in the inference stage. 
\paragraph{Inference.}
Data for evaluating our model is sampled from nuScenes validation dataset. For evaluation, we sample two evaluation subsets consisting of clips starting with nuScenes keyframes. The first dataset samples all valid clips and involves 5369 samples in total, where 'valid' means clips have enough frames for condition and acting as generation ground truth (number of frames at least 32). The second dataset samples 4 clips for each scene out of all 150 scenes with an interval of 59 frames, with the clips' format consistent with the first dataset. We generate video tokens for 50 steps with our temporal progressive denoising scheduler with a fixed classifier-free guidance scale~\cite{classifier-free-guidance} of $3.0$. The text template we used is as follow.
\begin{lstlisting}[language=Java]
This is a driving scene video captured by
camera in {sensor} of a car at {location}
during {season} {daytime}.
\end{lstlisting}

\begin{table}[t]
    \centering
    \footnotesize
    \begin{tabular}{c|cccc}
        \hline
         \multirow{2}{*}{Strategy} & Background & Subject & Motion & Temporal \\
         &Consistency & Consistency & Smoothness & Flickering \\
         \hline
         DC & 91.44\% & \textcolor{red}{90.02\%} & 97.42\% & 95.86\% \\
         CR & \textcolor{red}{91.75\%} & 89.93\% & \textcolor{red}{97.64\%} & \textcolor{red}{96.35\%} \\
         \hline
    \end{tabular}
    \caption{Video smoothness comparison~\cite{vbench}, where DC represents Divide-and-Conquer and CR represents Coarse-to-(Re)fine. \textcolor{red}{Red} represent the best value.}
    \label{tab:ablation_study_corr}
    \vspace{-5mm}
\end{table}
\vspace{-3mm}
\begin{figure}[b]
    \centering
    \includegraphics[width=\linewidth]{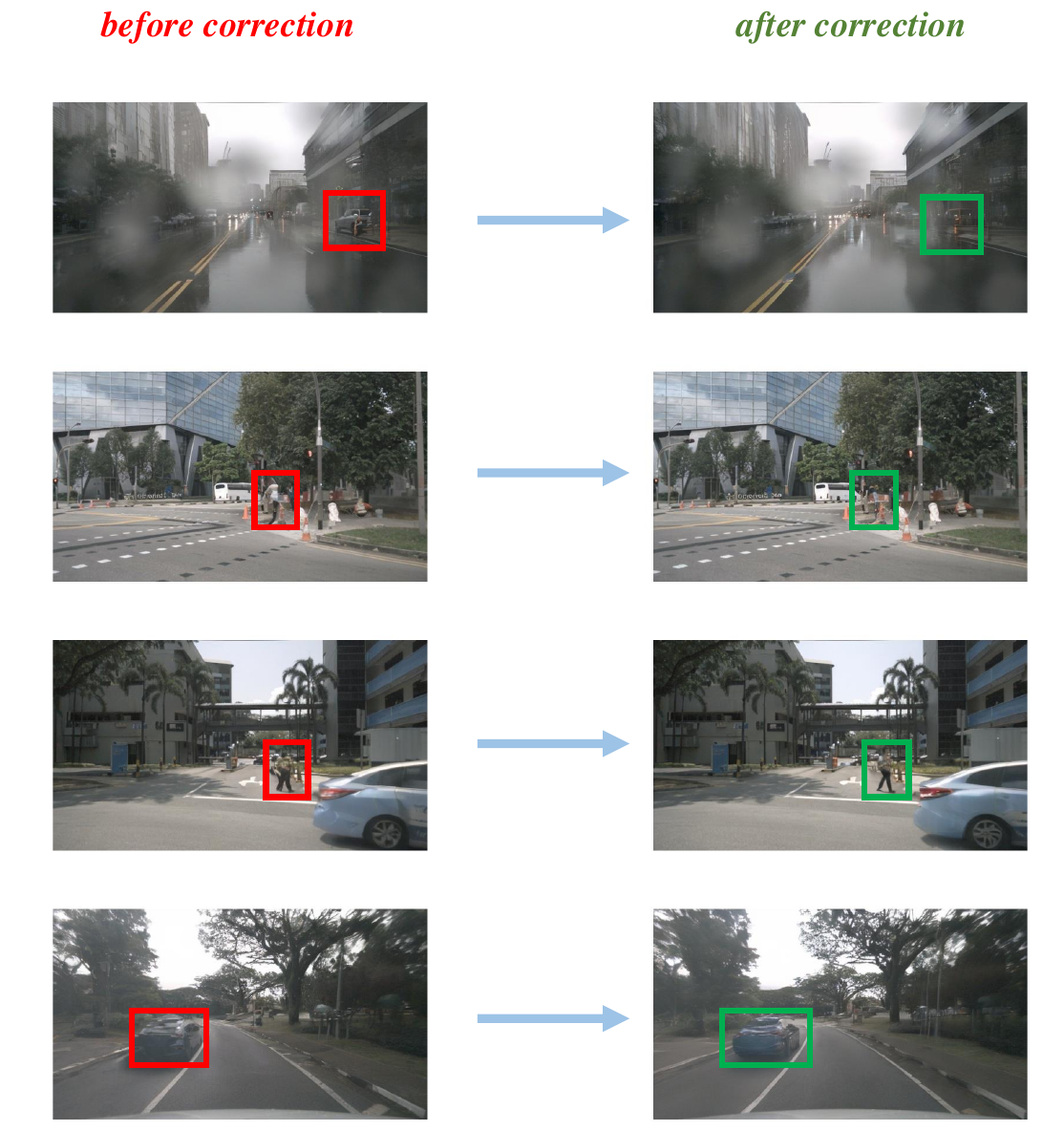}
    \caption{Visualization of anchor frame correction.}
    \label{fig:anchor_frame_correction}
\end{figure}
\section{Additional Results}
\begin{figure*}[!htbt]
    \centering
    \includegraphics[width=\textwidth]{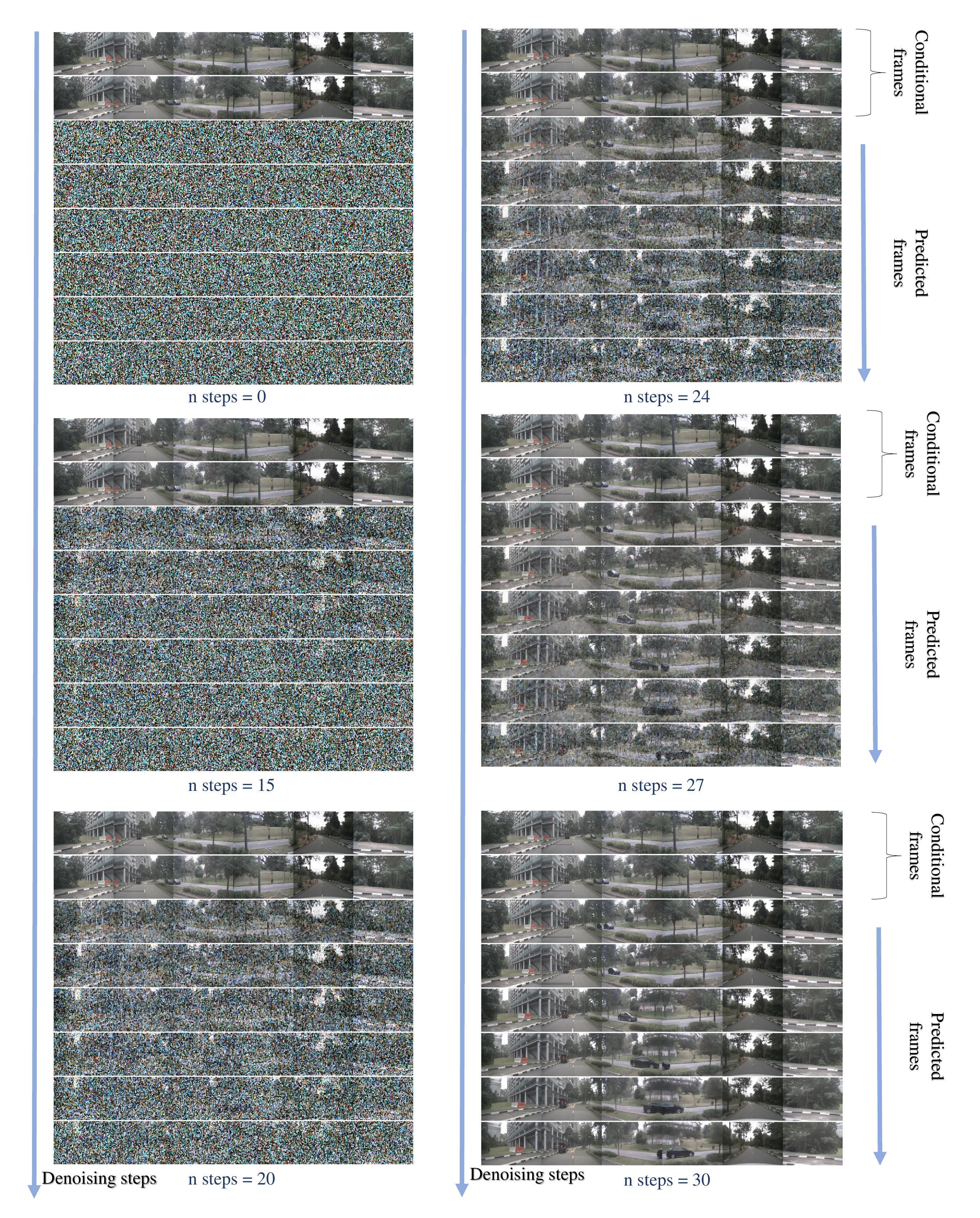}
    \caption{Denoising steps with various Predicted frames index. Frames closer to the conditional frames are earlier denoised as in Sec. \textcolor{cyan}{3.3}.}
    \label{fig:denoise}
\end{figure*}

\subsection{Additional Results for Progressive Denoising}
The Denoising steps with various Predicted frame indexes are illustrated in Fig.~\ref{fig:denoise}.  Visualisation in Fig.~\ref{fig:noise_scale_comp} offers additional insights into the fidelity of the generated videos. The frames produced using temporal progressive denoising successfully capture more detailed information about the scene.
\subsection{Additional Results for Trajectories’ Guidance}
We have conducted experiments to assess the influence of trajectories’ guidance, where results are recorded in Table.~\ref{tab:ablation_study_traj}. The results demonstrated the impact of trajectories’ guidance would be stronger for longer videos. More specifically, when examining cases with 12 frames, we found that FID and FVD scores for both front-view and multiple-view models were compatible with or without trajectory data. However, when analyzing cases with 24 frames, the models utilizing trajectory data significantly outperformed those without. These findings provide evidence that our MiLA is capable of generating videos that adhere to specific trajectory instructions.
\subsection{Additional Results for Long Video Generation}
\paragraph{Versus Recurrent Generation}
\begin{figure*}[!htbt]
    \centering
    \includegraphics[width=\textwidth]{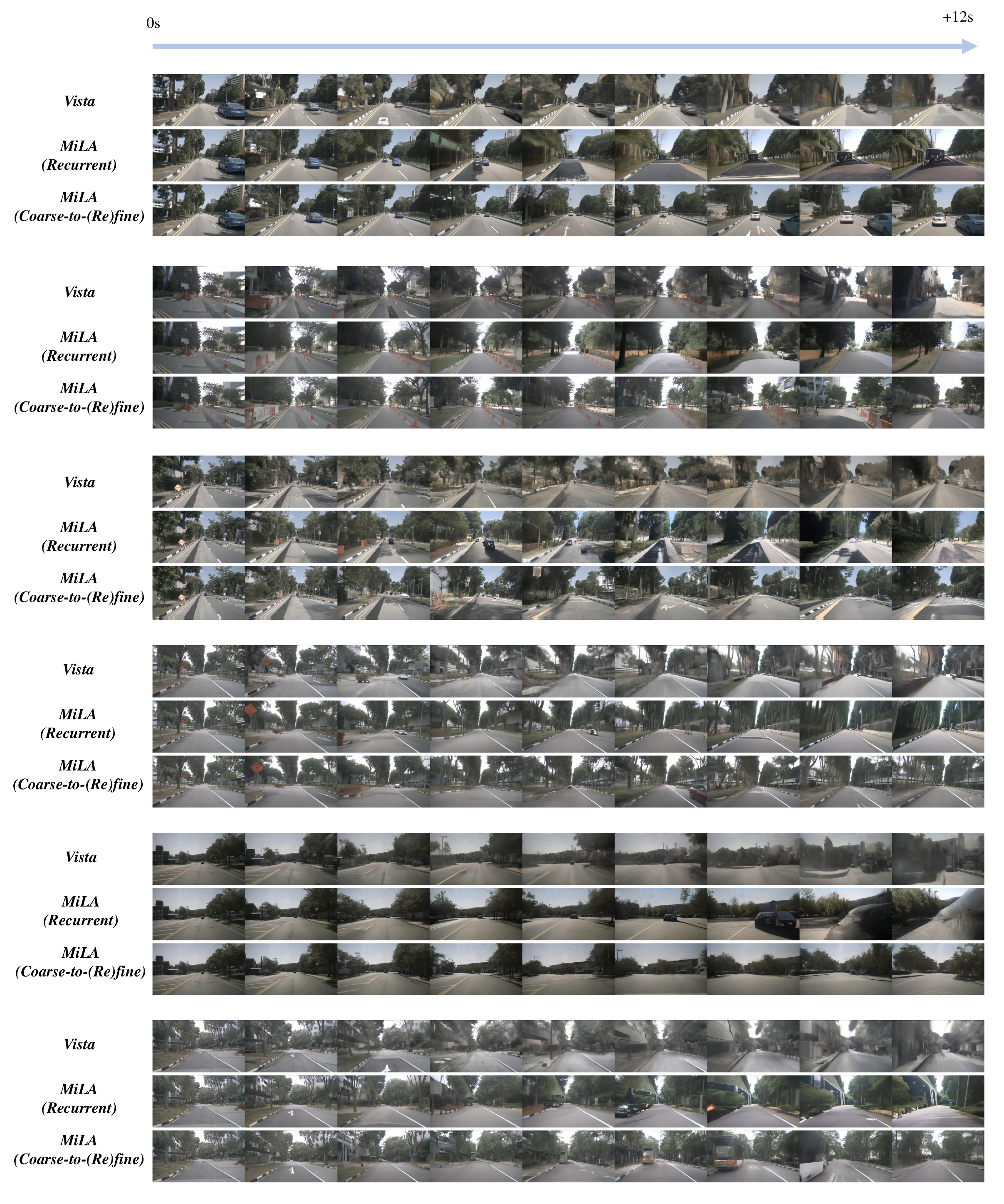}
    \caption{Visualization of 12-second 12-FPS video generation. It is noteworthy that MiLA with the Coarse-to-(Re)fine strategy maintains consistency in most dynamic scenes, while other Recurrent-based models gradually generate distorted images.}
    \label{fig:12s_compare}
\end{figure*}
We have conducted extensive experiments across Recurrent generation (including Vista and the Mila with Recurrent generation) and the proposed Coarse-to-(Re)fine framework and some additional qualitative results are illustrated in Fig.~\ref{fig:12s_compare}. We also provide an additional AVI file that captures the comparison between the MiLA (Coarse-to-(Re)fine) and Vista. To conduct a thorough analysis of the videos generated with the Vista and the MiLA, videos with 12 seconds are predicted, and only the front-view frames are utilized for comparison. As illustrated, most videos generated by Vista and MiLA (Recurrent) tend to distort and predict out-of-distribution frames after 10 seconds for dynamic scenes. However, the MiLA (Coarse-to-(Re)fine) maintains fidelity throughout the entire generation process. 

\paragraph{Versus Divide-and-Conquer Generation}
In this section, we will analyze the effectiveness of the correction process of the MiLA as shown in Fig.~\ref{fig:anchor_frame_correction}. After the correction process, objects in the red boxes (left col) are fixed (green boxes, right col). To this end, the Consistency of the videos is improved, see Table.~\ref{tab:ablation_study_corr}. The Coarse-to-(Re) framework achieves better performance in Consistency and smoothness compared to the Divide-and-Conquer framework.

\subsection{Additional Visualization Results}
\begin{figure*}[!htbt]
    \centering
    \includegraphics[width=\textwidth]{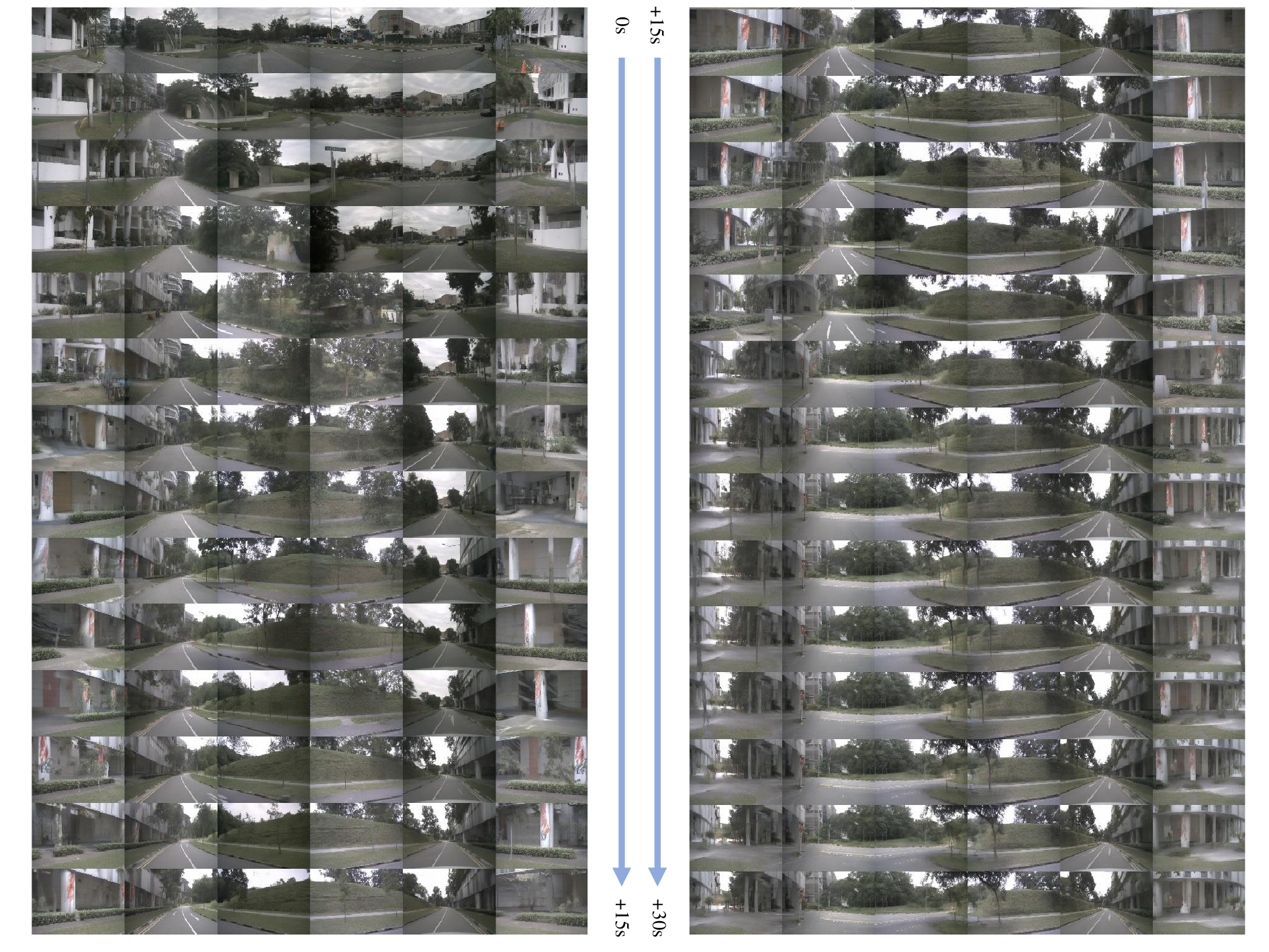}
    \caption{A comprehensive demonstration of the MiLA's video generation capabilities is presented, highlighting a 30-second clip produced at 12 FPS and featuring a detailed analysis of each sampled frame. The demonstration records and samples one frame per second.}
    \label{fig:super_long}
\end{figure*}

\begin{figure*}[!htbt]
    \centering
    \includegraphics[width=\textwidth]{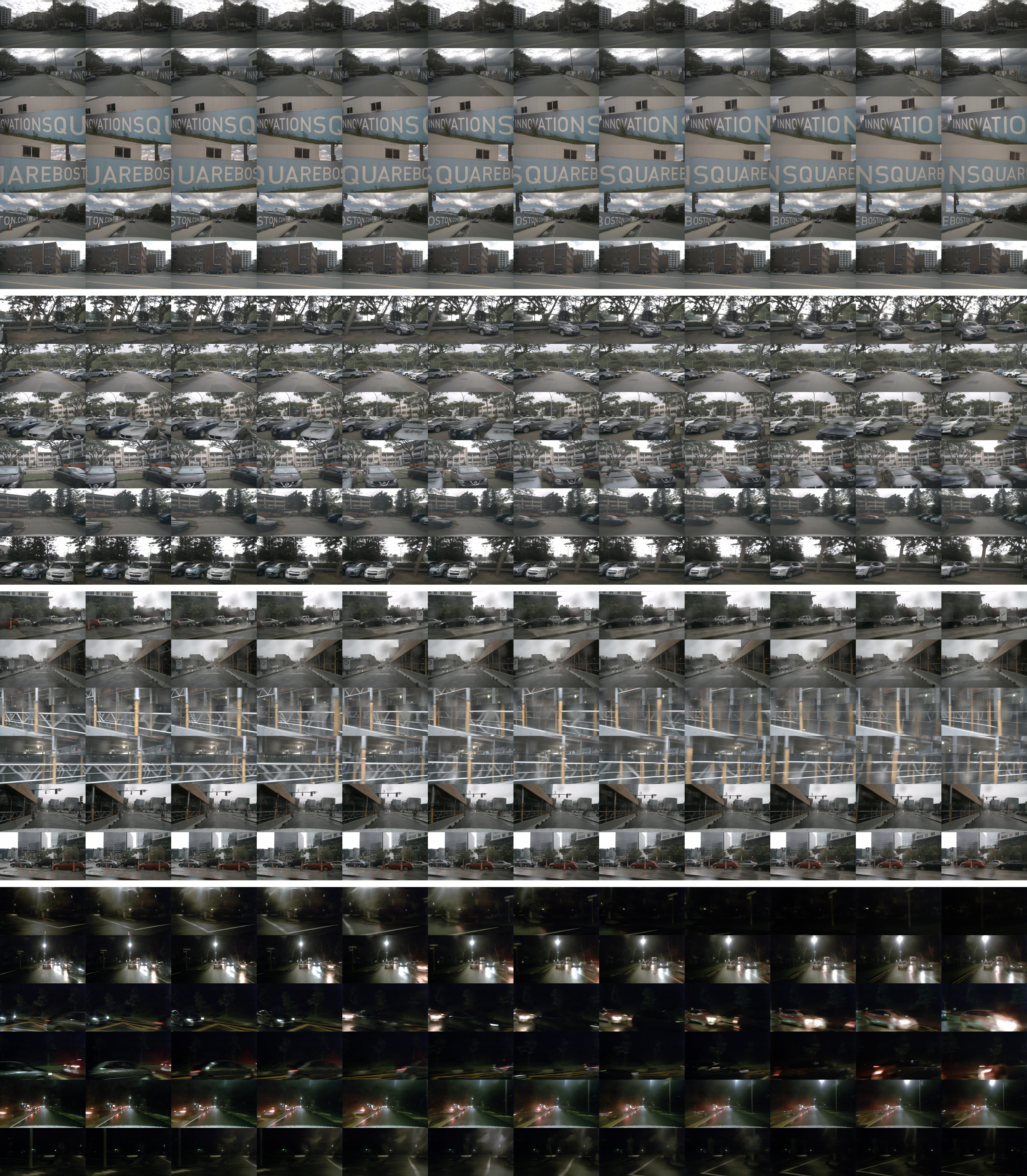}
    \caption{Visualization of MiLA's 12-FPS short-term video generation conditioned on 4 frames. MiLA achieves impressive performance on various hard cases. All frames demonstrated in this Figure are predicted by the MiLA.}
    \label{fig:short_vis}
\end{figure*}

A 12-FPS long-term video clip is shown in~\ref{fig:super_long}. We also provide an additional AVI file that captures a long-term video produced by the MiLA along with the supplementary material. These demonstrations serve as evidence of the MiLA's capability to generate long-term videos. Furthermore, various videos with different scenes are shown in Fig.~\ref{fig:short_vis}. The selected scenes incorporate hard samples with 
complicated street scenes, numerous vehicles, and quick cars in the darkness, \emph{et al.}  In conclusion, this evidence proves the MiLA can generate high-fidelity videos in most hard cases.

\subsection{Additional Results of Downstream Model Recognition}

As discussed in previous works ~\cite{genad, drivewm}, video quality alone is insufficient for a comprehensive evaluation of generated driving scene videos. In addition to visual quality, it is crucial to incorporate 3D layout perception or road segmentation techniques to assess the structural differences between generated and ground truth videos, particularly in reconstruction tasks. Unfortunately, to date, no existing method provides code for evaluating such structural similarities.


In this work, we evaluate the 3D layouts of videos generated by MiLA using a VAD model ~\cite{vad} trained on the nuScenes training set. This is done by replacing all image inputs with those generated by MiLA. As shown in Table \ref{tab:odld_metric}, our results demonstrate that MiLA can generate frames that exhibit both compatibility and recognizability, even in the absence of explicit 3D structural information during generation.

\begin{table}[h]
    \vspace{-3mm}
    \centering
    \small
    \begin{tabular}{c|cccc}
        \hline
        {\small Method} & {\small $\text{mAP}_{OD}$} & {\small $\text{NDS}_{OD}$} & {\small $\text{mAP}_{Line}$} & {\small $\text{Recall}_{Line}$}\\
         \hline
        GT & 0.326 & 0.454 & 0.465 & 0.655\\
        MiLA & 0.241 & 0.388 & 0.388 & 0.592\\
         \hline
    \end{tabular}
    \vspace{-3mm}
    \caption{Generation controllability.}
    \vspace{-3mm}
    \label{tab:odld_metric}
\end{table}

\end{document}